# Minority Oversampling for Imbalanced Time Series Classification

Tuanfei Zhu[1], Cheng Luo[3], Jing Li[1], Siqi Ren[4] and Zhihong Zhang[2,*]

[1]College of Computer Engineering and Applied Mathematics, Changsha University, Changsha, 410000, China

[2]School of Information Technology and Management, Hunan University of Finance and Economics, Changsha, 410205, China

[3]School of Information Engineering, Jiangxi Vocational Technical College of Industry and Trade, Nanchang, 330038, China

[4]School of Computer and Information Engineering, Zhejiang Gongshang University, Hangzhou, 310000, China

[*]Corresponding Author: Zhihong Zhang. Email: zhzhang@ccsu.edu.cn

**Abstract:** Many important real-world applications involve time-series data with skewed distribution. Compared to conventional imbalance learning problems, the classification of imbalanced time-series data is more challenging due to high dimensionality and high inter-variable correlation. This paper proposes a structure preserving Oversampling method to combat the High-dimensional Imbalanced Time-series classification (OHIT). OHIT first leverages a density- ratio based shared nearest neighbor clustering algorithm to capture the modes of minority class in high-dimensional space. It then for each mode applies the shrinkage technique of large-dimensional covariance matrix to obtain accurate and reliable covariance structure. Finally, OHIT generates the structure-preserving synthetic samples based on multivariate Gaussian distribution by using the estimated covariance matrices. Experimental results on several publicly available time-series datasets (including unimodal and multimodal) demonstrate the superiority of OHIT against the state-of-the-art oversampling algorithms in terms of F1, G-mean, and AUC.

## 1 Introduction

In this paper, we focus our attention on oversampling techniques in data-level approaches, since oversampling directly addresses the difficulty source of classifying imbalanced data by compensating the insufficiency of minority class information, and, unlike undersampling, does not suffer the risk of discarding informative majority samples.

## 2 The Proposed OHIT Framework

**Algorithm 1** OHIT($P$, $k$, $\kappa$, $drT$, $\eta$)

**Require:** *P*: the minority sample set; $k$, $\kappa$, $drT$ : three parameters in DRSNN clustering; $\eta$: the number of synthetic samples required to be generated;

**Ensure:** *Syn*: the generated synthetic sample set

1: Employ DRSNN to cluster the minority class sample set, $C_i \leftarrow$ DRSNN($P, k, \kappa, drT$), $i = 1, 2, ...m$, where $m$ is the number of discovered clusters.

2: Compute the shrinkage covariance matrix $\mathbf{S}_i^*$ for each cluster $C_i$ by combining Eqns. 7, 14, and 15.

3: Generate the synthetic sample set $Syn_i$ with size $\left\lfloor \eta \frac{|C_i|}{|P|} \right\rfloor$ for each cluster $C_i$ based on $N(\mu_i, \mathbf{S}_i^*)$, then add $Syn_i$ into *Syn*.

## 3 Experimental Results and Discussion

**Table 1:** Summary of the imbalanced unimodal time-series datasets used in the experiments

| Dataset | Minority Class | Length | Training data | | Testing data | |
|---|---|---|---|---|---|---|
| | | | Class Distribution | IR | Class Distribution | IR |
| Yoga (Yg) | '1' | 426 | 137/163 | 1.19 | 1393/1607 | 1.15 |
| Herring (Hr) | '2' | 512 | 25/39 | 1.56 | 26/38 | 1.46 |
| Strawberry (Sb) | '1' | 235 | 132/238 | 1.8 | 219/394 | 1.8 |
| PhalangesOutlinesCorrect (POC) | '0' | 80 | 628/1172 | 1.87 | 332/526 | 1.58 |
| Lighting2 (Lt2) | '-1' | 637 | 20/40 | 2 | 28/33 | 1.18 |
| ProximalPhalanxOutlineCorrect (PPOC) | '0' | 80 | 194/406 | 2.09 | 92/199 | 2.16 |
| ECG200 (E200) | '-1' | 96 | 31/69 | 2.23 | 36/64 | 1.78 |
| Earthquakes (Eq) | '0' | 512 | 35/104 | 2.97 | 58/264 | 4.55 |
| Two_Patterns (Tp) | '2' | 128 | 237/763 | 3.22 | 1011/2989 | 2.96 |
| Car | '3' | 577 | 11/49 | 4.45 | 19/41 | 2.16 |
| ProximalPhalanxOutlineAgeGroup (PPOA) | '1' | 80 | 72/328 | 4.56 | 17/188 | 11.06 |
| Wafer (Wf) | '-1' | 152 | 97/903 | 9.3 | 665/5499 | 8.27 |

*IR is the imbalance ratio (#majority class samples/#minority class samples).*

**Table 2:** Summary of the imbalanced multimodal time-series datasets used in the experiments

| Dataset | Minority class | Length | Training data | | Testing data | |
|---|---|---|---|---|---|---|
| | | | Class distribution | IR | Class distribution | IR |
| Worms (Ws) | '5','2','3' | 900 | 31/46 | 1.48 | 73/108 | 1.48 |
| Plane (Pl) | '3','5' | 144 | 36/69 | 1.92 | 54/51 | .944 |
| Haptics (Ht) | '1','5' | 1092 | 51/104 | 2.04 | 127/181 | 1.43 |
| FISH | '4','5' | 463 | 43/132 | 3.07 | 57/118 | 2.07 |
| UWaveGestureLibraryAll (UWGLA) | '8','3' | 945 | 206/690 | 3.35 | 914/2668 | 2.92 |
| InsectWingbeatSound (IWS) | '1','2' | 256 | 40/180 | 4.5 | 360/1620 | 4.5 |
| Cricket_Z (CZ) | '3','5' | 300 | 52/338 | 6.5 | 78/312 | 4 |
| SwedishLeaf (SL) | '10','7' | 128 | 54/446 | 8.26 | 96/529 | 5.51 |
| FaceAll (FA) | '1','2' | 131 | 80/480 | 12 | 210/1480 | 7.05 |
| MedicalImages (MI) | '5','6','8' | 99 | 23/358 | 15.57 | 69/691 | 10 |
| ShapesAll (SA) | '1','2','3' | 512 | 30/570 | 19 | 30/570 | 19 |
| NonInvasiveFatalECG_Thorax1 (NIFT) | '1','23' | 750 | 71/1729 | 24.35 | 100/1865 | 18.65 |

*IR is the imbalance ratio (#majority class samples/#minority class samples).*

**Table 3:** Performance results of all the compared methods on the imbalanced unimodal datasets

| Metrics | Methods | Datasets | | | | | | | | | | |
|---|---|---|---|---|---|---|---|---|---|---|---|---|
| | | Yg | Hr | Sb | POC | Lt2 | PPOC | E200 | Eq | TP | Car | PPOA | Wf |
| F1 | NONE | .565 | NaN | **.957** | .358 | .578 | .701 | .714 | NaN | .578 | .688 | .424 | .316 |
| | ROS | .607 | .405 | .951 | **.564** | .612 | .738 | .743 | .111 | .637 | .813 | **.532** | .651 |
| | SMOTE | .595 | **.503** | *.937* | .481 | **.696** | .724 | .768 | .195 | .638 | **.840** | .437 | .600 |
| | MDO | .607 | .406 | .948 | .554 | .630 | .747 | .734 | *.062* | .628 | .688 | .496 | .600 |
| | INOS | **.614** | .417 | .949 | .506 | .604 | .743 | .754 | .075 | .647 | .781 | .505 | .615 |
| | MoGT | .572 | *.392* | .945 | .535 | .653 | .743 | **.774** | .187 | .592 | .688 | .436 | **.684** |
| | OHIT | .613 | .466 | .951 | .550 | .654 | **.750** | .765 | **.197** | **.649** | .827 | .510 | .561 |
| G-mean | NONE | .618 | *.000* | .969 | .475 | .639 | .751 | .773 | .000 | .682 | .742 | .626 | .437 |
| | ROS | .639 | .509 | .968 | **.635** | .662 | .813 | .800 | .266 | .772 | .858 | .769 | **.830** |
| | SMOTE | .623 | **.582** | *.960* | .568 | **.722** | .806 | .822 | .382 | **.786** | **.882** | **.831** | .810 |
| | MDO | .640 | .511 | .965 | .630 | .673 | .814 | .790 | .184 | .759 | .742 | .726 | .813 |
| | INOS | **.645** | .521 | .967 | .596 | .656 | .818 | .809 | .203 | .779 | .826 | .776 | .811 |
| | MoGT | *.607* | .499 | .964 | .616 | .689 | .818 | **.826** | .379 | .742 | .742 | .718 | .813 |
| | OHIT | .642 | .558 | **.969** | .626 | .695 | **.822** | .818 | **.396** | .783 | .869 | .793 | .808 |
| AUC | NONE | .677 | *.249* | .990 | .669 | .706 | **.904** | .903 | .468 | .850 | .929 | .891 | .802 |
| | ROS | .677 | .617 | .991 | .669 | *.699* | *.884* | .894 | .532 | .853 | **.936** | *.873* | **.885** |
| | SMOTE | .662 | **.634** | .993 | *.631* | .722 | .886 | .900 | **.566** | .856 | .931 | .901 | .760 |
| | MDO | .677 | .603 | .992 | **.675** | **.727** | .899 | .896 | .531 | .856 | .926 | .900 | .875 |
| | INOS | **.686** | .621 | .992 | .651 | .700 | .888 | *.892* | .532 | **.863** | .931 | .908 | .863 |
| | MoGT | *.637* | .624 | *.989* | .670 | .701 | .888 | **.913** | .527 | *.820* | *.907* | .896 | *.758* |
| | OHIT | .682 | .627 | .992 | .665 | .706 | .899 | .901 | .534 | .860 | .934 | **.910** | .871 |

*Best (/Worst) results are highlighted in bold (/italics) Type.*

**Table 4:** Performance results of all the compared methods on the imbalanced multimodal datasets

| Metrics | Methods | Datasets | | | | | | | | | | | |
|---|---|---|---|---|---|---|---|---|---|---|---|---|---|
| | | Ws | Pl | Ht | FISH | UWGLA | IWS | CZ | SL | FA | MI | SA | NIFT |
| F1 | NONE | *.027* | .962 | *.439* | .863 | .718 | .618 | NaN | *.367* | .826 | *.200* | .667 | **.735** |
| | ROS | .464 | .967 | .537 | **.900** | *.653* | *.585* | .370 | **.812** | **.834** | .351 | .478 | .695 |
| | SMOTE | .505 | **.980** | .622 | .875 | .705 | .660 | **.553** | .639 | .810 | .393 | *.475* | *.354* |
| | MDO | .493 | .972 | .608 | .884 | .676 | .675 | .431 | .733 | .819 | **.610** | .684 | .646 |
| | INOS | .468 | .968 | .592 | .896 | .735 | .685 | .473 | .786 | .815 | .436 | .476 | .680 |
| | MoGT | .468 | .975 | .615 | .888 | .728 | **.696** | *.390* | .787 | *.756* | .443 | .620 | .587 |

|  | OHIT | **.505** | .967 | **.629** | .893 | **.741** | .683 | .515 | .787 | .811 | .449 | **.687** | .693 |
|---|---|---|---|---|---|---|---|---|---|---|---|---|---|
|  | NONE | *.117* | .962 | *.539* | .875 | .793 | .708 | *.000* | .478 | .877 | *.358* | .729 | .804 |
|  | ROS | .549 | .967 | .612 | **.922** | .767 | .772 | .530 | .878 | .886 | .633 | .867 | .807 |
|  | SMOTE | .568 | **.980** | .672 | .918 | .836 | .844 | **.754** | .875 | **.904** | .756 | .911 | **.879** |
| G-mean | MDO | .572 | .972 | .667 | .900 | .786 | .798 | .599 | .848 | .888 | .773 | .930 | *.786* |
|  | INOS | .553 | .968 | .654 | .916 | .842 | .847 | .642 | .890 | .901 | .766 | .900 | .837 |
|  | MoGT | .553 | .975 | .673 | .905 | .829 | .841 | .576 | .905 | .887 | .739 | .908 | .803 |
|  | OHIT | **.577** | .967 | **.684** | .913 | **.845** | **.850** | .691 | **.907** | .900 | **.780** | **.931** | .852 |
|  | NONE | *.436* | .998 | .710 | .950 | .907 | .884 | .735 | .959 | .960 | .856 | .927 | .961 |
|  | ROS | .555 | *.996* | *.674* | .949 | *.863* | *.839* | .721 | .951 | **.961** | *.712* | *.911* | .971 |
|  | SMOTE | **.574** | .998 | .713 | .950 | .909 | .898 | **.822** | .944 | .953 | .798 | .924 | *.953* |
| AUC | MDO | .559 | .999 | .711 | .946 | .877 | .875 | .760 | *.921* | .954 | .805 | .933 | .968 |
|  | INOS | .565 | *.996* | .708 | **.952** | .907 | .901 | .774 | .949 | .953 | .855 | .921 | .972 |
|  | MoGT | .560 | .997 | .711 | *.945* | .910 | .896 | *.695* | .939 | *.939* | .838 | .927 | *.956* |
|  | OHIT | .571 | **.999** | **.729** | .951 | **.911** | **.901** | .814 | **.961** | .954 | **.877** | **.936** | **.973** |

*Best (/Worst) results are highlighted in bold (/italics) Type.*

**Table 5:** Summary of *p*-values of Wilcoxon significance tests between OHIT and each of the other compared methods

| OHIT vs | Unimodal data | | | Multimodal data | | |
|---|---|---|---|---|---|---|
|  | F1 | G-mean | AUC | F1 | G-mean | AUC |
| Original | $9.8e-4_+$ | $9.8e-4_+$ | $0.0552_*$ | $0.0122_+$ | $4.9e-4_+$ | $0.0044_+$ |
| ROS | 0.4131 | $0.0342_+$ | $0.0425_+$ | $0.042_+$ | $0.0015_+$ | $0.0034_+$ |
| SMOTE | 0.6772 | .9097 | .3013 | $0.0342_+$ | 0.6221 | $0.0342_+$ |
| MDO | $0.0342_+$ | $0.0093_+$ | .377 | .1099 | $0.0015_+$ | $0.0024_+$ |
| INOS | $0.0342_+$ | $0.0049_+$ | $0.021_+$ | $0.0425_+$ | $0.0068_+$ | $0.0015_+$ |
| MoGT | $0.064_*$ | $0.0122_+$ | $0.0269_+$ | $0.0269_+$ | $0.0015_+$ | $4.9e-4_+$ |

**Table 6:** Recall, specificity, and precision of SMOTE and OHIT on the imbalanced unimodal datasets

| Metrics | Methods | Datasets | | | | | | | | | | |
|---|---|---|---|---|---|---|---|---|---|---|---|---|
|  |  | Yg | Hr | Sb | POC | Lt2 | PPOC | E200 | Eq | TP | Car | PPOA | Wf |
| Recall | SMOTE | .590 | **.477** | **.991** | .486 | .679 | **.808** | **.864** | .167 | **.868** | .842 | **.847** | .713 |
|  | OHIT | **.605** | .404 | .985 | **.559** | .589 | .796 | .806 | **.185** | .771 | .816 | .694 | **.732** |
| Specificity | SMOTE | .658 | .713 | .930 | .664 | .770 | .805 | .783 | **.879** | .712 | .924 | .816 | **.919** |
|  | OHIT | **.682** | **.774** | **.952** | **.703** | **.821** | **.849** | **.831** | .849 | **.796** | **.927** | **.907** | .893 |
| Precision | SMOTE | .600 | .533 | .888 | .477 | .716 | .658 | .692 | **.235** | .505 | **.838** | .295 | **.518** |
|  | OHIT | **.622** | **.551** | **.920** | **.542** | **.737** | **.709** | **.729** | .212 | **.561** | .838 | **.403** | .455 |

*In terms of recall, specificity and precision, p-values of Wilcoxon test between OHIT and SMOTE are 0.1763, 0.0161, and 0.0674, respectively.*

**Table 7:** Average performance of OHIT and its variants across all the datasets within each group

| OHIT vs | Unimodal data | | | Multimodal data | | |
|---|---|---|---|---|---|---|
|  | F1 | G-mean | AUC | F1 | G-mean | AUC |
| OHIT/DRSNN | .6229 | .7290 | .7935 | .6641 | .8131 | .8751 |
| OHIT/shrinkage | .5988 | .6977 | .7938 | .6523 | .7769 | .8475 |
| OHIT with ER | .5938 | .6974 | .7939 | .6619 | .7900 | .8519 |
| OHIT | **.6243** | **.7316** | **.7984** | **.6966** | **.8247** | **.8813** |

## 4 Conclusion

Extensive experiments have been conducted to evaluate the effectiveness of OHIT on both the unimodal datasets and multimodal datasets. In most of case, OHIT can significantly outperform existing representative oversampling solutions in terms of F1, G-mean, and AUC (Table 5).

**Acknowledgement:** We thank the authors of MDO, INOS, and MoGT for sharing their algorithm codes with us.

Funding Statement**:** This work was supported in part by the National Natural Science Foundation of China (Grant NO: 62006030, 61906167), in part by the National Natural Science Foundation of Hunan Province, China (Grant NO: 2020JJ5623, 2020JJ4648), in part by the Scientific Research Foundation of Hunan Provincial Education Department (Grant NO: 20B059, 20B060, 19A048).

**Conflicts of Interest:** The authors declare that they have no conflicts of interest to report regarding the present study.